\documentclass{article}

\usepackage{arxiv}
\usepackage[utf8]{inputenc} 
\usepackage[T1]{fontenc}  
\usepackage{hyperref}      
\usepackage{url}           
\usepackage{booktabs}       
\usepackage{amsfonts}       
\usepackage{nicefrac}       
\usepackage{microtype}     
\usepackage{lipsum}
\usepackage{times}
\usepackage{epsfig}
\usepackage{graphicx}
\usepackage{amsmath}
\usepackage{amssymb}
\usepackage{diagbox}
\usepackage{subcaption}
\usepackage{lineno}
\usepackage{biblatex}
\usepackage{algorithm} 
\usepackage{algpseudocode} 
\usepackage{amssymb}
\usepackage{caption} 
\usepackage[flushleft]{threeparttable}
\usepackage{setspace}
\usepackage{wrapfig}
\usepackage[dvipsnames]{xcolor}
\usepackage{bm}
\usepackage{pdflscape}

\title{Open-set learning with augmented categories by exploiting unlabelled data}

\author{
  Emile-Reyn Engelbrecht \\
  Main and corresponding author \\
  Electronic Engineering \\
  Stellenbosch University \\
  South Africa \\
  \texttt{18174310@sun.ac.za} \\
   \And
  Johan A. du Preez \\
  Co-author \\
  Electronic Engineering \\
  Stellenbosch University \\
  South Africa \\
  \texttt{dupreez@sun.ac.za} \\
}

\begin{document}
\maketitle
\begin{abstract}
Novel categories are commonly defined as those unobserved during training but present during testing. However, partially labelled training datasets can contain unlabelled training samples that belong to novel categories, meaning these can be present in training and testing. This research is the first to generalise between what we call observed-novel and unobserved-novel categories within a new learning policy called open-set learning with augmented category by exploiting unlabelled data or Open-LACU. After surveying existing learning policies, we introduce Open-LACU as a unified policy of positive and unlabelled learning, semi-supervised learning and open-set recognition. Subsequently, we develop the first Open-LACU model using an algorithmic training process of the relevant research fields. The proposed Open-LACU classifier achieves state-of-the-art and first-of-its-kind results.
\end{abstract}

% keywords can be removed

%labelling vs annotation
%Classifier vs model

\section{Introduction}
\paragraph{}
Inductive reasoning is the process of generalising principles to align observations with their respective outcomes. In the context of classification, inductive models learn about inputs that belong to distinct source categories in order to classify new inputs into one of these categories. Recently, substantial research efforts have been concentrated on inductive classification using highly-parameterised neural networks~\cite{lecun2015deep}. These networks have consistently achieved state-of-the-art results across various machine learning tasks, often surpassing human performance~\cite{he2015delving, esteva2017dermatologist}. Despite these achievements, the popular learning policies used to train neural networks are resource-intensive and fail to encompass all category types found in application-grade datasets. 

Supervised learning (SL) is the most common inductive machine learning policy used to train neural network classifiers~\cite{goodfellow_deep_learning}. However, SL is hindered by 1) cost, since it demands a large number of labelled training samples~\cite{van2020survey}; and 2) practicality, since it requires every category in the domain to have a corresponding output label~\cite{engelbrecht2020learning}. To mitigate (1), semi-supervised learning (SSL) is a well-researched alternative that trains classifiers using partially labelled datasets or labelled and unlabelled training data. SSL is capable of training high-accuracy classifiers using considerably fewer labelled training samples than SL, with the assumption of a large number of unlabelled training samples. However, SSL is also burdened by the second constraint, that is, it assumes a closed set for the training-testing pipeline.

A closed set implies that all training and testing samples correspond to one of the pre-defined training labels, which define the source categories. However, gathering labels for every category represented in raw unlabelled data is exceptionally costly, and so it is common for some of the unlabelled training data to also belong to novel categories~\cite{bekker2020learning}. Furthermore, data patterns are known to change over time, meaning new novel categories can emerge post-training~\cite{geng2020recent, gruhl2021novelty}. Formally, a novel category is a group of anomalous samples that, as a group, exhibit similar patterns, but these patterns do not match those any of those source categories~\cite{gruhl2021novelty}. To ensure safe classifications, samples from all novel categories must be detected and separated from the samples that belong to the source categories of interest. 

Novelty detection research generally focuses on emerging novel categories, meaning they were \mbox{\textit{unobserved}} during training but presented during testing~\cite{pimentel2014review, geng2020recent}. A critical example of an unobserved-novel category is the SARS-CoV-2 virus prior to its discovery~\cite{andersen2020proximal}. Formally, unobserved-novel categories only exist in the future domain and so cannot be represented during training, i.e. the present. Therefore, separating unobserved-novel categories is essential for both accurate classifications and to provide practitioners the chance to study the new evolving patterns~\cite{gruhl2021novelty}. However, unobserved-novel categories do not account for all novel patterns encountered in the training-testing pipeline. Instead, novel categories that have data samples \mbox{\textit{observed}} within unlabelled training data can also exist in the partially labelled scenarios~\cite{blanchard2010semi, da2014learning, bekker2020learning, yu2020multi, guo2020safe}. These observed-novel categories are especially prevalent in big data applications wherein the cost to annotate every category in the training set is extremely high~\cite{engelbrecht2020learning}.

The timing and location of novel categories within the training-testing pipeline varies across different learning policies. Here, we note that although some policies focus solely on one-class novelty detection, we study the policies that consider multi-class classification with simultaneous novelty detection. Therefore, the fields of concern are open-set recognition (OSR), which is the policy that defines novel categories as those unobserved during training but encountered during testing~\cite{geng2020recent, pimentel2014review}, and learning with augmented category by exploiting unlabelled data (LACU)~\cite{da2014learning}, which is the policy that defines novel categories as those observed in unlabelled training data. However, no previous learning policy has acknowledged and effectively addressed both observed-novel and unobserved-novel within a single training scheme. Instead, most studies consider one or the other, while some studies suggest that observed-novel categories are equivalent to unobserved-novel categories. Considering the temporal differences between these novel category types, we argue that future models must generalise them separately. 

This study presents the unified machine learning policy of open-set learning with augmented category by exploiting unlabelled data or Open-LACU. Formally, Open-LACU is a combination of OSR and LACU, noting that LACU is really a combination of positive and unlabelled (PU)-learning and SSL. Open-LACU models are trained using partially labelled datasets that contain samples from source categories in the labelled and unlabelled training data and samples from observed-novel categories scattered within the unlabelled training data. Subsequently, Open-LACU models are presented with testing samples from the source categories, the observed-novel categories, and unobserved-novel categories that did not have any data represented during training. Formally, Open-LACU models are expected to generalise 1) $K > 1$ number of source categories (as per general supervised classification), 2) an additional augmented $K + 1$'th \textit{background} category wherein all samples from observed-novel categories must be classified, and 3) an augmented $K + 2$'nd \textit{unknown} category wherein all samples from unobserved-novel categories must be classified. Thus, Open-LACU classifiers are the first capable of classification with simultaneous observed and unobserved novelty detection. 

Our proposed Open-LACU model divides the training process into two distinct steps. First, we train a one-class PU-learning model with the objective of distinguishing between the source categories placed into the augmented positive category and the observed-novel categories. This model then acts as a filter to identify some samples that belong to the observed-novel categories. Post PU-training, the identified samples are labelled into the augmented $K + 1$'th \textit{background} category which are combined with the original labelled training data to train the classifier network. However, considering the abundance of unlabelled training samples that remain available (i.e. those that were not filtered), an SSL method is used to train the classifier networks, but with $K + 1$ nodes in its softmax. Finally, we evaluate the trained classifiers' performance in accurately classifying all $K + 1$ categories and its ability to detect samples from unobserved-novel categories using the classic reject option approach.

Although several difficulties arise within our proposed procedure, our method is only presented as a prototype for the Open-LACU policy. In other words, several additional factors should be considered in future research, yet our model holds state-of-the-art and first-of-its-kind results. The remainder of this study is structured as follows: Section~2 provides background information on classification-based learning policies, with a specific focus on the different types of categories within each policy; Section~3 introduces the unified learning policy of Open-LACU, focusing on the necessary inference methodology for such a classifier and the algorithmic training-testing procedure; Section~4 discusses our proposed model(s), Section~5 conducts various experiments to prove its effect; Section 6~discusses the results and presents future considerations for Open-LACU; and Section~7 concludes the study.
 
\section{Background}
\label{Chap52}
Conventional semi-supervised learning (SSL) focuses on the inductive setting wherein classifiers are trained on a partially labelled training set, \mbox{$\mathcal{D}_{\text{train}} := \mathcal{D}_{\text{lab-train}} \cup \mathcal{D}_{\text{unlab-train}}$}, and tested on an independent testing set, $\mathcal{D}_{\text{test}}$~\cite{van2020survey}. These sets contain labelled pairs $(x,y)$, where $x$ is the input sample and $y$ is the corresponding output label. However, only labels from $\mathcal{D}_{\text{lab-train}}$ are available to the classifier, which defines the source categories, while labels in $\mathcal{D}_{\text{unlab-train}}$ and $\mathcal{D}_{\text{test}}$ are unavailable or, in the spirit of research, hidden from the classifier. In general SSL, samples in $\mathcal{D}_{\text{unlab-train}}$ and $\mathcal{D}_{\text{test}}$ are assumed to belong to one of the source categories, and so we refer to the labels in these sets as anticipated labels, $y_a$. Again, it is crucial to note that conventional SSL does not account for any novel categories within its training-testing pipeline. Let \mbox{$C_K := \{1, 2, ..., K\}$} denote the unique labels for each of the $K$ source categories. Then, SSL has all \mbox{$\{y \sim \mathcal{D}_{\text{lab-train}}\} \in C_K$}, all \mbox{$\{y_a \sim \mathcal{D}_{\text{unlab-train}}\} \in C_K$}, and all \mbox{$\{y_a \sim \mathcal{D}_{\text{test}}\} \in C_K$}.

In the scenarios where some categories are novel and so do not have corresponding labels, $\mathcal{D}_{\text{unlab-train}}$ and/or $\mathcal{D}_{\text{test}}$ could contain samples from both source categories represented in $C_K$ and any number of novel categories not represented in $C_K$~\cite{guo2020safe, geng2020recent}. Novel categories can be handled in of two ways: 1) in class-incremental learning (CIL) or zero-shot learning, each novel category must be appended to $C_K$ with a unique label as it is discovered, after which the classifier must generalise that category~\cite{van2022three, rizve2022openldn}, or 2) in novelty detection, novel categories need only be identified and placed into a separate category (e.g. the background or unknown category)\cite{bekker2020learning, geng2020recent}. Intuitively, novelty detection serves as a natural precursor to achieving CIL, and this study only focuses on the former. However, there exist two distinct types of novel categories in literature - observed-novel and unobserved-novel categories.

Observed-novel categories have data samples observed in a section of the unlabelled training set, $\mathcal{D}_{\text{unlab-train}}$. Conventional SSL methods have been shown to falter in the presence of observed-novel categories~\cite{oliver2018realistic, willetts2020semi, banitalebi2022auxmix}. As a result, several alternative SSL research branches have emerged to mitigate performance degradation which include 1) safe/auxiliary/mismatched-SSL, which seeks to reduce the negative impact of observed novel categories on SSL models~\cite{guo2020safe, banitalebi2022auxmix, chen2020semi}, and 2) universal SSL, which extends mismatched-SSL to the case of domain adaptation~\cite{huang2021universal}. However, these research branches do not explicitly require of trained classifiers to detect observed-novel categories. In other words, samples from observed-novel categories are not included within $\mathcal{D}_{\text{test}}$. Given that our aim is novelty detection, novel categories would certainly exist in the testing phase. 

To the best of our knowledge, only two research fields consider classification with simultaneous observed-novelty detection. These fields are open-set semi-supervised learning (open-SSL)~\cite{yu2020multi, luo2021consistency, huang2021trash, saito2021openmatch} and learning with augmented category by exploiting unlabelled data (LACU)~\cite{da2014learning, schrunner2020generative, pham2016semi, engelbrecht2020learning}. We also acknowledge the field of open-world SSL~\cite{rizve2022openldn} or semi-unsupervised learning~\cite{willetts2020semi}, which extends open-SSL and LACU to the CIL setting. Disregarding CIL, let all observed-novel categories be given the $K + 1$'th label. Open-SSL and LACU are then formally defined by a training-testing pipeline that has all \mbox{$\{y \sim \mathcal{D}_{\text{lab-train}}\} \in C_K$}, all \mbox{$\{y_a \sim \mathcal{D}_{\text{unlab-train}}\} \in \{C_K \cup K + 1\}$}, and all \mbox{$\{y_a \sim \mathcal{D}_{\text{test}}\} \in \{C_K \cup K + 1\}$}. However, here it is essential to discuss the conflated definitions of observed-novel and unobserved-novel categories within existing research. 

Open-SSL considers observed-novel and unobserved-novel categories as one and the same. However, open-set recognition (OSR), which open-SSL is a sub-branch of, defines novel categories as those that are unobserved during training but seen during testing~\cite{pimentel2014review, geng2020recent}. Here, let all unobserved-novel categories be given the $K + 2$ category label. Although OSR studies generally train classifiers under a supervised setting (i.e. using a fully-labelled dataset), our recent study showed how OSR benefits from a SSL setting (which we simply refer to as SSL-OSR)~\cite{engelbrecht2023link}. Given the additional $K + 2$ label, SSL-OSR is defined with a training-testing pipeline that has all \mbox{$\{y \sim \mathcal{D}_{\text{lab-train}}\} \in C_K$}, all \mbox{$\{y \sim \mathcal{D}_{\text{unlab-train}}\} \in C_K$} and all \mbox{$\{y_a \sim \mathcal{D}_{\text{test}}\} \in \{C_K \cup K + 2\}$}. It is clear that SSL-OSR does not include observed-novel categories within its pipeline. Consequently, the assumption of open-SSL that observed-novel and unobserved-novel categories are one and the same does not adhere to the definitions set out within literature. 

In the next section, we will show how the conflated definitions of observed-novel and unobserved-novel categories in open-SSL leads to algorithmic failure in trying to distinguish between the two. Here, it is only important to note that LACU does not conflate these definitions, and so extending LACU to OSR provides means of detecting both novel category types. Finally, it is important to note the relationship between positive and unlabelled (PU)-learning and LACU. PU-learning is the LACU scenario where $K = 1$~\cite{bekker2020learning}. Therefore, LACU can be considered a combination of general SSL and PU-learning, which is important in relation to our proposed Open-LACU training procedure. Finally, as a summary, Table~\ref{5table:1} indicates the category types found in the training-testing pipelines of all the above discussed research branches. It is clear that none but Open-LACU, formally proposed in the next section, includes all novel category types in its training-testing pipeline. 

\begin{landscape}
    \begin{table*}[t!]
        \centering
        \begin{tabular}{m{5.0cm} m{1.5cm} m{1.5cm} m{2.5cm} m{1.5cm} m{1.5cm} m{1.5cm}}
        \hline
         \multicolumn{1}{c}{} & \multicolumn{2}{c}{\textbf{Category types in training}} & \multicolumn{1}{c}{} & \multicolumn{3}{c}{\textbf{Category types in testing}} \\
        \hline
        & Source & Observed-novel & & Source & Observed-novel & Unobserved-novel \\
        \hline
        SL & \multicolumn{1}{c}{\checkmark} & & & \multicolumn{1}{c}{\checkmark} & & \\
        OSR &  \multicolumn{1}{c}{\checkmark} & & & \multicolumn{1}{c}{\checkmark} & & \multicolumn{1}{c}{\checkmark}  \\
        \hline
        SSL & \multicolumn{1}{c}{\checkmark} & & & \multicolumn{1}{c}{\checkmark} & &  \\
        SSL-OSR & \multicolumn{1}{c}{\checkmark} & & & \multicolumn{1}{c}{\checkmark} & & \multicolumn{1}{c}{\checkmark} \\
        Mismatched-SSL & \multicolumn{1}{c}{\checkmark} & \multicolumn{1}{c}{\checkmark} & & \multicolumn{1}{c}{\checkmark} & &  \\
        PU-learning ($K = 1$) & \multicolumn{1}{c}{\checkmark} & \multicolumn{1}{c}{\checkmark} & & \multicolumn{1}{c}{\checkmark} & \multicolumn{1}{c}{\checkmark} &  \\
        Open-SSL / LACU & \multicolumn{1}{c}{\checkmark} & \multicolumn{1}{c}{\checkmark} & & \multicolumn{1}{c}{\checkmark} & \multicolumn{1}{c}{\checkmark} &  \\
        Open-LACU & \multicolumn{1}{c}{\checkmark} & \multicolumn{1}{c}{\checkmark} & & \multicolumn{1}{c}{\checkmark} & \multicolumn{1}{c}{\checkmark} & \multicolumn{1}{c}{\checkmark}   \\
        \hline
        \end{tabular}
        \caption{Category types found in the training-testing pipeline of classification based machine learning policies. Unless otherwise indicated, all learning policies study multi-category classification (i.e. $K > 1$). Learning policies concerned with class-incremental learning and domain adaption are not included in this table. The policies are split into two groups: fully labelled training sets (top two) and partially labelled training sets (bottom six).}
        \subcaption*{\\ \textbf{Key}: SL is supervised learning, OSR is open-set recognition, SSL is semi-supervised learning, PU-learning is positive and unlabelled learning, open-SSL is open-set semi-supervised learning, LACU is learning with augmented category by exploiting unlabelled data, and Open-LACU is our proposed policy of Open-set LACU.}
        \label{5table:1}
    \end{table*}
\end{landscape}

\section{Open-LACU}
No prior learning policy has simultaneously considered source categories, observed-novel categories, and unobserved-novel categories. Open-set recognition (OSR) considers source categories and unobserved-novel categories, while learning with augmented categories using unlabelled data (LACU) considers source categories and observed-novel categories. Therefore, a unified approach of OSR and LACU would encompass all relevant category types. This section formally introduces this new learning policy of Open-LACU, specifically focusing on the training-testing pipeline, the required inference method of Open-LACU classifiers, and the algorithmic process to train and test an Open-LACU classifier. 

\subsection{Training-testing criteria}
Open-LACU classifiers generalise $K$ number of source categories, an augmented $K + 1$'th background category that encapsulates observed-novel categories, and an augmented $K + 2$'nd unknown category that encapsulates unobserved-novel categories. Using the previous section's notation, Open-LACU has all \mbox{$\{y \sim \mathcal{D}_{\text{lab-train}}\} \in C_K$}, all \mbox{$\{y_a \sim \mathcal{D}_{\text{unlab-train}}\} \in \{C_K \cup K + 1\}$}, and all \mbox{$\{y_a \sim \mathcal{D}_{\text{test}}\} \in \{C_K \cup K + 1, K + 2\}$}. It is important to reiterate that there are no labelled training samples available for the $K + 1$ and $K + 2$ categories since these represent the two novel category types. However, unlabelled training samples are available for the augmented $K + 1$'th background category, noting that these are scattered in an amongst unlabelled training samples from the $K$ source categories. 

\subsection{Inference}
\begin{table*}[t!]
   \small
   \centering
   \begin{tabular}{m{3cm} m{3cm} m{1.0cm} m{4cm}}
   \toprule\toprule
   \textbf{Novelty detector 1} & \textbf{Novelty detector 2} & & \textbf{Predicted category} \\ 
   \midrule
   $d_1(x) > \tau_1$ & $d_2(x) > \tau_2$ & $\longrightarrow$ & $p_c(x)$ \\
   $d_1(x) < \tau_1$ & $d_2(x) > \tau_2$ & $\longrightarrow$ & $ K + 1$ \\
   $d_1(x) > \tau_1$ & $d_2(x) < \tau_2$ & $\longrightarrow$ & $K + 2$ \\
   $d_1(x) < \tau_1$ & $d_2(x) < \tau_2$ & $\longrightarrow$ & $K + 1$ or $K + 2$ \\
   \bottomrule
   \end{tabular}
    \caption{Ambiguity in classifications when using two reject options for observed-novel and unobserved-novel categories} 
   \subcaption*{\\ \textbf{Key}: $d_1(x)$ is the probability score of the observed-novelty detector given input sample $x$; $d_2(x)$ is the probability score of the unobserved-novelty detector given input sample $x$; $\tau_1$ is the upper-limit threshold applied to $d_1(x)$; $\tau_2$ is the upper-limit threshold applied to $d_2(x)$, and $p_c(x)$ is the predicted category for input sample $x$ across $K$ source categories.}
   \label{5table:2}
\end{table*}

Novelty detection is commonly applied with a reject option on the output of the classifier or one-class novelty detector. A reject option is a threshold applied to the predicted probabilities of a classifier, and any sample with a prediction that does not meet the threshold is deemed from a novel category. OSR generally applies a reject option to the output of the multi-class classifier to detect unobserved-novel categories~\cite{geng2020recent, pimentel2014review}. However, open-SSL also applies a reject option, albeit on a one-class novelty detector, to detect observed-novel categories~\cite{yu2020multi}. Open-LACU requires separation of observed-novel and unobserved-novel categories, and so we cannot make use of reject options for both. More specifically, two reject options would develop ambiguity in the classification of the $K + 1$'th and $K + 2$'nd categories. 

Two reject options demand two separate networks for the observed-novel and the unobserved-novel categories, respectively. Let $d_1(x)$ represent the probability score of the observed-novelty detector given input sample $x$, and let $d_2(x)$ represent the probability score of the unobserved-novelty detector. Whether $d_1(x)$ and/or $d_2(x)$ is a one-class classifier or multi-class classifier is irrelevant for this discussion. Then, let $\tau_1$ represent the upper-limit rejection threshold value for $d_1$, and let $\tau_2$ represent the upper-limit rejection threshold value for $d_2$. Consequently, any sample with an predicted probability below these thresholds is detected the designated novel category as per the model. However, ambiguity is developed in using two reject options, as shown by the different events in Table~\ref{5table:2}. 

The first option in Table~\ref{5table:1} states that neither of the reject options were activated, and so the sample can be classified accordingly into one of the $K$ source categories with the maximum prediction (represented as $p_c(x)$). The next two options activate one or the other reject options, and so the input sample is deemed from an observed-novel or unobserved-novel category depending on which reject option is activated. However, the final option activates both reject options, and so now it is uncertain whether the input sample belongs to an observed-novel category or an unobserved-novel category. Therefore, we argue that the unnecessary computational complexity of having two detectors with reject options and the ambiguity that proceeds such a setup motivate the use of a LACU network. Specifically, a LACU network appends the $K + 1$'th category to the softmax output of the classifier network, alleviating the need for a reject option to detect observed-novel categories. 

In LACU, the augmented $K + 1$'th background category is appended to the output of the neural network as a separate label to encapsulate all observed-novel categories. Although the first neural-LACU method used a threshold on the $K + 1$'th output node~\cite{engelbrecht2020learning}, we will later show that recent advancements in PU-learning have made it possible to generalise the augmented category under the same softmax activation function as the $K$ source categories. In other words, a classifier network is able to classify a sample into any of the $K$ source categories or into the $K + 1$ augmented category under a single softmax. Consequently, we can achieve classification with simultaneous observed and unobserved novelty detection by applying a reject option on a LACU classifier with $K + 1$ labels. Therefore, Open-LACU does not fall into the trap of ambiguity when using two reject options. 

With respect to inference, a LACU classifier produces a probability score across $K + 1$ categories, denoted as \mbox{$C(x) = [C_1(x), C_2(x), ..., C_{K - 1}(x), C_K(x), C_{K + 1}(x)]$}. Each entry in $C(x)$ is computed using the softmax activation function, represented as \mbox{$C_i(x) = \frac{\exp{(c_i(x)})}{\sum_{k = 1}^{K + 1}\exp{(c_k(x))}}$}, where $c_i(x)$ is the network logit at position $i$ prior to the softmax. Subsequently, the classifier's predicted category for input sample $x$ is determined as \mbox{$p_c(x) = \text{max}\big[C(x)\big]$}. However, for the detection of unobserved-novel categories, we apply a reject option on $\text{max}\big[C(x)\big]$ with a specified threshold $\tau$. Any sample with $p_c(x)  < \tau$ is classified into the $K + 2$ category, indicating that $x$ belongs to an unobserved-novel category. Formally, the inference process for an Open-LACU classifier is as follows:

\begin{equation}
    p_c(x) = 
        \begin{cases}
            \max[C(x)] & \text{if } \max[C(x)] > \tau \\
            K + 2 & \text{otherwise}
        \end{cases}
\end{equation}

\paragraph{}

\subsection{Algorithmic procedure}
Open-LACU represents a combination of OSR and LACU. However, as discussed in the Background (Chapter~\ref{Chap52}), LACU is a combination of PU-learning and SSL. Therefore, we propose an algorithmic process that draws from these different research areas to train an Open-LACU model. Initially, it is essential to recognise that a classifier with $K + 1$ output nodes cannot effectively generalise to the $K + 1$'th category without any labelled training samples for that category. Therefore, we first train a PU-learning model to filter out at least some of the samples that belong to observed-novel categories. These filtered samples are then labelled into the $K + 1$'th category and be appended with the original labelled training samples from the other $K$ source categories to train a $K + 1$ classifier. Considering that unlabelled training samples are still available, we opt for SSL to train this classifier. Finally, in the testing phase, unobserved-novel categories are introduced to facilitate OSR in the mix. 

Consider a PU-learning model with parameters $\theta_1$, which is trained by minimising any general PU-learning loss function, $\mathcal{L}_{\text{PU}}(\mathcal{D}_{\text{lab-train}}, \mathcal{D}_{\text{unlab-train}})$. Specifically, $\mathcal{L}_{\text{PU}}$ updates $\theta_1$ in a such way that it maximises the separation between source categories and observed-novel categories. PU models generally employ a one-class classifier that produces a single probability output score denoted as $d_{\theta_1}(x)$. To extend our classifier to accommodate the augmented $K + 1$'th background category, we utilise the trained $d_{\theta_1}(x)$ to create a new dataset, $\mathcal{D}_{\text{obs}}$, which consists of some samples that belong to observed-novel categories. Specifically, we apply a threshold, $\tau_{\text{PU}}$, to the PU model, effectively treating it as a one-class classifier with reject option, to create \mbox{$\mathcal{D}_{\text{obs}} := \{\forall x \sim \mathcal{D}_{\text{unlab-train}} \; \; | \; \; d_{\theta_1}(x) < \tau_{\text{PU}}\}$}.

It is crucial to emphasize that $\mathcal{D}_{\text{obs}}$ must encompass all observed-novel categories present in the training data. If $\mathcal{D}_{\text{obs}}$ only represents a subset of observed-novel categories, the classifier might not effectively generalise the augmented $K + 1$'th background category. Assuming that $\mathcal{D}_{\text{obs}}$ includes some samples from all observed-novel categories, these samples are provided with the $K + 1$'th label before proceeding to train the classifier model. For the classifier, we train an SSL model with parameters $C_{\theta_2}$, which minimises any general SSL loss function $\mathcal{L}_{\text{SSL}}$. However, we append a supervised loss to $\mathcal{L}_{\text{SSL}}$ for the $K + 1$'th labelled samples in $\mathcal{D}_{\text{obs}}$. It is also worth noting that the PU model and the classifier model can potentially share parameters, i.e. $\theta_1 = \theta_2$. This concept is called multi-task learning~\cite{tanaka2018joint}, but we leave development of such a framework for future studies.

The loss function $\mathcal{L}_{\text{SSL}}$ can take the form of any SSL loss function defined over $\mathcal{D}_{\text{lab-train}}$ and $\mathcal{D}_{\text{unlab-train}}$. To incorporate the newly formed set $\mathcal{D}_{\text{obs}}$ into the training framework, we can simply append a supervised loss term for the samples in $\mathcal{D}_{\text{obs}}$ that have been labelled as part of the $K + 1$'th category: \mbox{$\mathcal{L}_{\text{LACU}} = \mathcal{L}_{\text{SSL}} -  \mathbb{E}_{(x) \sim D_{\text{obs}}} [\log(C_{K + 1}(x))]$}, where $C_{K + 1}(x)$ represents the output of the classifier at the $K + 1$'th node. It is important to note that this LACU training procedure remains agnostic to the specific PU-learning and SSL loss functions. However, the algorithmic training procedure of these two policies is necessary to ensure that the classifier generalises over the $K$ source training categories and the augmented $K + 1$'th background category for observed-novel categories.

Lastly, the trained LACU classifier undergoes testing under the LACU and OSR policy (i.e. Open-LACU). Specifically, the testing criteria must encompass samples from the $K$ source categories, samples from all observed-novel categories ($K + 1$), and samples from unobserved-novel categories ($K + 2$). Given that the $K + 1$'th category can encompass numerous observed-novel categories, it is common for this category to contain more testing samples than the source categories. Therefore, LACU generally employs the macro-F1 score to evaluate the classifier's performance, which removes bias for imbalanced testing data. Subsequently, OSR typically utilises the area under the receiver operating curve (AUROC) to gauge the ability to detect unobserved-novel categories using a reject option. Thus, the Open-LACU testing phase returns two metrics, the macro-F1 over $K + 1$ labels and the AUROC for the $K + 2$'nd label.

\begin{algorithm}[t!]
    \setstretch{1.2} % Increase line spacing by 20%
    \caption{Training and testing}
    \label{5algo:open-lacu}
    \begin{algorithmic}[1]
        \Procedure{Open-LACU}{}
            \State \textbf{Input:} $\mathcal{D}_{\text{lab-train}}, \mathcal{D}_{\text{unlab-train}}, \tau_{\text{PU}}, \mathcal{D}_{\text{test}}$, $d_{\theta_1}(x)$, $C_{\theta_2}$
            \State \textit{PU-learning:}
            \State \hspace{1em} Train $d_{\theta_1}(x)$ by minimisng $\mathcal{L}_{\text{PU}}(\mathcal{D}_{\text{lab-train}}, \mathcal{D}_{\text{unlab-train}}$) until convergence.
            \State \hspace{1em} Develop $\mathcal{D}_{\text{obs}} := \{\forall x \sim \mathcal{D}_{\text{unlab-train}} \; \; | \; \; d_{\theta_1}(x) < \tau_{\text{PU}}\}$
            \State \textit{SSL:}
            \State \hspace{1em} Train $C_{\theta_2}$ by minimising $\mathcal{L}_{\text{LACU}} = \mathcal{L}_{\text{SSL}} - \; \mathbb{E}_{(x) \sim D_{\text{obs}}} [\log(c_{K + 1}(x))]$
            \State \textit{Open-LACU testing:}
            \State \hspace{1em} Evaluate $C_{\theta_2}$ using $\mathcal{D}_{\text{test}}$ according to the macro-F1 and AUROC scores
            \State \textbf{Return:} Macro-F1, AUROC
        \EndProcedure
    \end{algorithmic}
\end{algorithm}

The complete algorithmic approach for training and testing an Open-LACU model is outlined in Algorithm~\ref{5algo:open-lacu}. Although the specific methods for the PU-model and the SSL classifier are not detailed here, following this step-by-step procedure is crucial for generalizing the an Open-LACU classifier. However, it is often considered more effective to streamline the various steps in training procedures. Not only does end-to-end loss functions reduce computational complexity but may also enhance the model's generalisation performance. Nevertheless, since this study serves as an introductory exploration of Open-LACU, the development of such end-to-end methods is left to future research. In this work, we only develop an Open-LACU classifier using the step-by-step method described above.

\section{Proposed method}
For the first Open-LACU model, we propose modifying the multi-task open-SSL model outlined by Yu et al.~\cite{yu2020multi}. It is essential to re-clarify that despite its name, open-SSL does not address the issue of unobserved-novel categories. Instead, Yu's approach centres around a one-class classifier with a reject option to identify observed-novel categories alongside a general SSL classifier for the source categories. In Yu's case, the one-class classifier is employed to separate samples that belong to observed-novel categories to ensure that the SSL classifier won't be negatively affected by these samples. In our model, we segment each process by YU to follow the Open-LACU procedure as described in Algorithm~\ref{5algo:open-lacu}.

First, we employ the one-class classifier to construct the $\mathcal{D}_{\text{obs}}$ dataset. Then, we augment the SSL classifier with an additional $K + 1$ output node and assign the samples in $\mathcal{D}_{\text{obs}}$ to the $K + 1$'th label. Subsequently, we proceed to train the SSL classifier with the appended supervised loss concerning the samples in $\mathcal{D}_{\text{obs}}$. Finally, we evaluate the performance of this classifier under the Open-LACU testing criteria. It is worth noting that the accuracy of developing $\mathcal{D}_{\text{obs}}$ holds significance for the classifier's training. Specifically, the PU-learning step should aim to attain two factors:~1) $\mathcal{D}_{\text{obs}}$ should encompass samples representing all observed-novel categories in the domain to ensure comprehensive generalisation for the augmented $K + 1$'th background category, and~2) $\mathcal{D}_{\text{obs}}$ should exclude any samples from the source categories as their presence would adversely affect the generalisation across the remaining $K$ labels.

While earlier models, particularly those in mismatched-SSL and open-SSL, may have tolerated some inaccuracies in their filtering process, Open-LACU models exhibit heightened sensitivity to this initial PU-learning phase. However, to the best of our knowledge, no literature currently describes the successful training of a PU-learning model capable of exact separation between source categories and observed-novel categories. Nevertheless, Yu et al. presented a step towards addressing this challenge through the enhanced adversarial binary cost function (E-ABC). The E-ABC function is an extension of the ABC function used for PU-learning in the first neural-LACU  model~\cite{engelbrecht2020learning}, and the PU-learning based GAN by Chiaroni et al.~\cite{chiaroni2020counter}. Given a one-class classifier $d(x)$, the ABC function is given as:

\begin{equation}
\mathcal{L}_{\text{ABC}} = - \; \; \mathbb{E}_{x \sim D_{\text{lab-train}}} \big[\log(1 - d(x)) \big] \; \; - \; \; \mathbb{E}_{x \sim D_{\text{unlab-train}}} \big[\log(d(x)) \big].
    \label{5eq:cost_abc}
\end{equation}

Under the ABC function, the PU-learning model is regarded as a one-class classifier which assigns a binary label of $1$ to all labelled training samples and a binary label of $0$ to all unlabelled training samples. As a result, an adversarial process is introduced since source categories contain samples with contrasting labels. Nevertheless, observed-novel categories exclusively consist of samples with the low binary label of $0$. Chiaroni et al.~\cite{chiaroni2020counter} demonstrated that $\mathcal{L}_{\text{ABC}}$ would train a model that consistently predicted $1$ for observed-novel categories, while it would generate outputs in the range of $0.5 < \delta < 1$ for all samples from source categories (noting a flip of labels in our case). However, this observation only holds true under the assumption of no overfitting on $ D_{\text{lab-train}}$.

Chiaroni designed a specific architecture to mitigate the risk of overfitting under the ABC function. However, one of the strengths of neural networks lies in their architectural flexibility. In the upcoming experiments section, we will demonstrate that a commonly used wide residual network or wide-ResNet tends to overfit on $D_{\text{lab-train}}$, and so $0.5 < \delta < 1$ becomes an approximation instead of a guaranteed solution. Consequently, the ABC function does not provide a suitable separation between source and observed-novel categories. Therefore, we present the E-ABC approach. For the E-ABC function, after pre-training the model using the general ABC function in eq.~\ref{5eq:cost_abc}, E-ABC alters the labels for all unlabelled training samples to follow the model's outputs. When considering $d_{t - 1}(x)$ as the model's output from the previous epoch for sample $x$, the E-ABC adjusts the ABC function after 10 epochs as follows:

\begin{equation}
\begin{split}
   \mathcal{L}_{\text{E-ABC}} = & - \; \; \mathbb{E}_{x \sim D_{\text{lab-train}}} \big[\log(1 - d(x)) \big] \\
   &  - \; \; d_{t - 1}(x) \; \cdot \; \mathbb{E}_{x \sim D_{\text{unlab-train}}} \big[\log(d(x)) \big] \\
   &  - \; \; (1 - d_{t - 1}(x)) \; \cdot \; \mathbb{E}_{x \sim D_{\text{unlab-train}}} \big[\log(1 - d(x)) \big]. \\
\end{split}
    \label{5eq:cost_eabc}
\end{equation}

We will later demonstrate that the E-ABC function offers a more effective separation between source and observed-novel categories. Therefore, with the PU-learning model trained using $\mathcal{L}_{\text{E-ABC}}$, the next step involves using the trained PU-learning model to filter out samples that belong to observed-novel categories. This entails selecting a threshold $\tau_{\text{PU}}$, as indicated in step 5 of Algorithm~\ref{5algo:open-lacu}. Since observed-novel categories were initially assigned the low label of $0$ for the PU-learning model, samples from these categories are expected to yield lower scores than the source categories. However, acknowledging that the PU-learning model is not flawless, an exact demarcation between these categories cannot be calculated and is not guaranteed. Nevertheless, our analysis confirms that the lowest output scores correspond to samples from observed-novel categories. Consequently, we set $\tau_{\text{PU}}$ as the low 10th percentile of all unlabelled training samples' outputs of the trained PU-learning model.

Here, it is essential to acknowledge that due to the imperfect nature of the PU-learning model, the created set, $\mathcal{D}_{\text{obs}}$, might inadvertently contain some samples from source categories. In the event that we mislabel these samples with the $K + 1$ label, we must train the classifier with the assumption of noisy labels. Although we found that such noisy labels did indeed exist, the E-ABC function ensures that $\mathcal{D}_{\text{obs}}$ always predominantly comprised of samples from observed-novel categories. Nevertheless, previous research showed that a high learning rate can negate the adverse effect of noisy labels by reducing overfitting~\cite{yu2020multi, tanaka2018joint}, and so we deploy a high learning rate for the SSL training step. It is important to reiterate that this first PU-learning step can greatly hinder model training, especially in the case where there are many observed-novel categories. Therefore, we advise that future research either focus on the noisy label problem or the development of a clean dataset for $\mathcal{D}_{\text{obs}}$ to ease the training procedure. 

Following PU-learning and the development of $\mathcal{D}_{\text{obs}}$, the subsequent step is to train an SSL classifier that is also capable of OSR. In an earlier work, we demonstrated the potential of generative SSL models for OSR~\cite{engelbrecht2023link}. However, due to the complexities associated with training generative adversarial networks (GANs), we could not extend the Open-LACU framework to incorporate GANs in this initial exploration of Open-LACU. Although the additional benefits GANs offer in SSL-OSR suggest that an Open-LACU GAN model could yield superior results. For example, a consolidated approach of the PU-learning based GAN by Chiaroni et al.~\cite{chiaroni2020counter} and the SSL-OSR GANs in our previous study~\cite{engelbrecht2023link} could greatly benefit Open-LACU. However, for our Open-LACU prototype, we exclusively employ the popular MixMatch SSL model~\cite{berthelot2019mixmatch} for our Open-LACU experiments and evaluate its performance in the context of OSR.

\section{Experiments}
Before presenting our Open-LACU results, it is imperative to establish that the E-ABC function effectively separates source categories from observed-novel categories. However, the experimental setup will remain consistent across all experiments, and so we describe these first. Specifically, we employ the MNIST and CIFAR10 datasets to emulate LACU training scenarios. We select $K = 2$, $K = 5$, or $K = 9$ as source categories, utilising the remaining $10 - K$ categories as observed-novel categories. For MNIST, in line with prior LACU experiments, we provide $1400$ labelled training samples for source categories and $4200$ unlabelled training samples for all source and observed-novel categories. In the case of CIFAR10, we provide $2000$ labelled training samples for source categories and $3000$ unlabelled training samples for all source and observed-novel categories. Then, during the testing phase, we append Fashion-MNIST to MNIST and CIFAR100 to CIFAR10 to create an OSR testing scenario.

\begin{figure*}[!t]
\centering
\centering
\subfloat[ABC]{\includegraphics[width=0.48\textwidth]{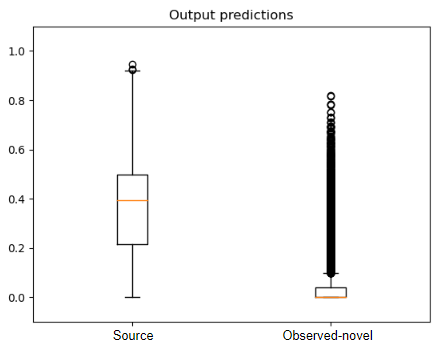}
\label{5fig:1a}}
\hfil
\centering
\subfloat[E-ABC]{\includegraphics[width=0.48\textwidth]{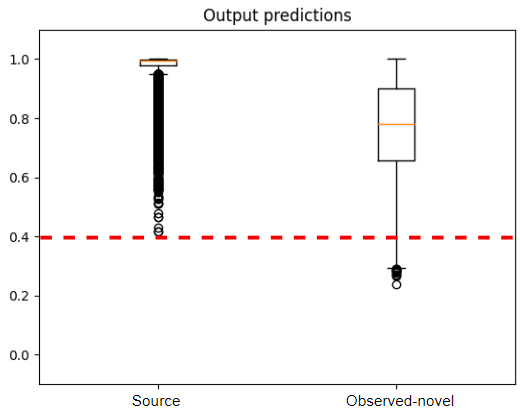}
\label{5fig:1b}}
\caption{Box and whisker plots indicating the separation of source and observed-novel categories for the CIFAR10 $K = 5$ experiment. The source box and whisker plot refers to the outputs of all training samples that belong to $K$ source categories, while the observed-novel refers to the output of all samples that belong to the $10 - K$ observed-novel categories. The red dotted line in (B) indicates the PU threshold to separate some samples from observed-novel categories, while (A) has no such available threshold.}
\label{5fig:1}
\end{figure*}

In our MNIST experiments, we employ the fully connected architecture utilised in our previous neural-LACU study~\cite{engelbrecht2020learning}. For CIFAR10, we opt for a wide-ResNet-28-2 architecture, indicating a depth of 28 and a width of 2. Concerning the separate PU-learning and SSL steps, we use the same architecture for both phases, yet we re-initialise the model for each step. Future research may explore a shared model for both steps, creating a multi-task framework. However, for the purpose of this study, we maintain separate models to facilitate a strict analysis of the two learning steps. Lastly, we uniformly set a large learning rate at $\text{lr} = 0.002$ for all models and experiments to address potential challenges posed by noisy labels.

Consider first the CIFAR10 $K = 5$ experiment for the PU-learning step. The box and whisker plots shown in Fig~\ref{5fig:1} depict the model outputs for both source and observed-novel categories after 200 epochs for the ABC and E-ABC functions, respectively. Although both loss functions exhibit some degree of separation, it is clear that the ABC function falls short in providing the mechanism to accurately filter out samples that belong to observed-novel categories, as required in step 5 of Algorithm~\ref{5algo:open-lacu}. To be more precise, there exists no distinct threshold, $\tau_{\text{PU}}$, which can be effectively applied to the model's predicted probabilities (see Fig.~\ref{5fig:1a}) to generate a clean dataset for $\mathcal{D}_{\text{obs}}$. In contrast, the E-ABC function can define such a threshold since it places samples from observed-novel categories behind those from source categories, as indicated by the red dotted line in Fig.~\ref{5fig:1b}.

\begin{table}[t]
    \centering
    %\begin{tabular}{m{3.2cm} m{2.2cm} m{2.2cm} m{0.3cm} m{2.2cm} m{2.2cm}}
    \begin{tabular}{ p{5.2cm} p{2.5cm} p{2.5cm} p{2.5cm}}
    \hline
     \multicolumn{4}{c}{\textbf{Open-LACU}}  \\
    \hline
        & MNIST & MNIST & MNIST \\
    Source categories ($K$) & 2 & 5  & 9  \\
    \hline
     Supervised & 75.99 $|$ 12.54 &  90.83 $|$ 23.07 &  93.18 $|$ 52.07 \\
     MixMatch & \textbf{97.78} $|$ \textbf{15.26} &  \textbf{97.33} $|$ \textbf{50.45} & \textbf{94.58}  $|$ \textbf{68.27} \\ 
     \multicolumn{4}{l}{} \\
     \textit{Our previous~\cite{engelbrecht2020learning}} &  \textit{80.02} $|$ - - - &  \textit{79.85} $|$ - - - & \textit{ 83.54}  $|$ - - - \\
     \textit{Schrunner~\cite{schrunner2020generative}} & \textit{79.00} $|$ - - - &  \textit{68.00} $|$ - - - &  \textit{25.00}  $|$ - - - \\
    \hline
    \multicolumn{4}{l}{\textit{Results are averaged over five trials}}
    \end{tabular} \\
    \caption{Best performer over the last 5 epochs for the Open-LACU MNIST experiments. Each result is split as $a \; | \; b $ with $a$ representing the \mbox{Macro-F1 score * $100$} over the $K + 1$ categories, and $b$ representing the \mbox{$\text{AUROC} * 100$}. The top-performers are shown in bold, and previously published results are in italics.}
    \label{5table:3}
\end{table}

Our preferred threshold for the E-ABC function is the bottom 10th percentile of the outputs from all unlabelled training samples (i.e. samples from source and observed-novel categories). This threshold ensures a fair choice to develop $\mathcal{D}_{\text{obs}}$ but can also induce a noisy label set. More specifically, unless a perfect PU-learning model is developed, it will be common for $\mathcal{D}_{\text{obs}}$ to contain correct samples from observed-novel categories and incorrect samples from source categories. Furthermore, it is essential to note that $\mathcal{D}_{\text{obs}}$ should contain samples from all $10 - K$ observed novel categories since it represents the augmented $K + 1$'th background category. However, this induces another layer of complexity since the label distribution might become heavily skewed towards the $K + 1$'th label. 

Nevertheless, we apply this PU-learning step after which we train a MixMatch SSL model (with the additional $K + 1$ labelled samples) for 100 epochs for the MNIST experiment and 1000 epochs for the CIFAR10 experiment. The results are laid out in Table~\ref{5table:3} and Table~\ref{5table:4}. As a comparison, we also trained a general supervised classifier over the $K + 1$ categories. For the MNIST experiments, it is immediately apparent that our proposed model achieves state-of-the-art results compared to previous LACU studies. However, it is worth noting that the AUROC scores from these models suggest that fully-connected networks are not effective for the separation of unobserved-novel categories. This conclusion can be confirmed with the far superior AUROC scores for the CIFAR10 experiments, which utilised a wide-ResNet architecture. This observation also challenges the recent assertion that a good closed-set classifier can seamlessly handle OSR~\cite{vaze2021open}, which was not the case for our MNIST experiments. Besides this, we also observe a noticeable uptick in AUROC scores as $K$ increases, a trend seen in both the MNIST and CIFAR10 experiments.

\begin{table}[t]
    \centering
    %\begin{tabular}{m{3.2cm} m{2.2cm} m{2.2cm} m{0.3cm} m{2.2cm} m{2.2cm}}
    \begin{tabular}{ p{5.2cm} p{2.5cm} p{2.5cm} p{2.5cm}}
    \hline
     \multicolumn{4}{c}{\textbf{Open-LACU}}  \\
    \hline
        & CIFAR10 & CIFAR10 & CIFAR10 \\
    Source categories ($K$) & 2 & 5  & 9  \\
    \hline
     Supervised & 21.93 $|$ 55.42 & 54.62 $|$ 65.91 &  80.47  $|$ 84.01 \\
     MixMatch & \textbf{72.33} $|$ \textbf{60.73} & \textbf{90.28} $|$ \textbf{77.81} & \textbf{84.429}  $|$ \textbf{84.22} \\
    \hline
    \multicolumn{4}{l}{\textit{Results are averaged over five trials}}
    \end{tabular} \\
    \caption{Best performer over the last 50 epochs for the Open-LACU CIFAR10 experiments. Each result is split as $a \; | \; b $ with $a$ representing the \mbox{Macro-F1 score * $100$} over the $K + 1$ categories, and $b$ representing the \mbox{$\text{AUROC} * 100$}. The top-performers are shown in bold.}
    \label{5table:4}
\end{table}

The increase in AUROC scores with the increase of $K$ suggests that neural networks grow more confident as they face the challenge of generalising a greater number of categories. This growing confidence manifests as higher prediction scores for in-domain samples and smaller scores for out-of-domain samples. However, we hypothesise that these shifts in scores are more likely attributed to the varying number of categories grouped in the augmented $K + 1$'th background category. Specifically, as $K$ increases, so $10 - K$ decreases which is the number of observed-novel categories within the domain. A decrease in observed-novel categories would result in a more confident model which also aligns well with our analysis of the PU-learning step. Specifically, the noisy label issue within $\mathcal{D}_{\text{obs}}$ would become less detrimental to the classifier if fewer categories were present within the augmented $K + 1$'th background category.

With regard to the supervised baselines, the imperfect development of $\mathcal{D}_{\text{obs}}$ also significantly hindered the training of these models. In particular, our experiments revealed that a supervised classifier couldn't effectively generalise the augmented $K + 1$'th background category, especially for the CIFAR10 scenarios. Hence, even with such a high number of labelled training samples, the supervised baselines always under performed. However, again we see an uptick in accuracy of the supervised classifier as $K$ increase. Clearly, further analysis is required to understand the impact of an imperfect $\mathcal{D}_{\text{obs}}$ and its representation of multiple $10 - K$ categories. Nevertheless, these results are state-of-the-art for general LACU and first-of-its-kind for Open-LACU. Thus, the first Open-LACU benchmark results are developed. 

\section{Discussion}
The Open-LACU problem is a complex process that involves navigating imperfect aspects of various research fields. When we examine the algorithmic process for training an Open-LACU model, these imperfections can compound and result in unsuccessful training. In particular, the imperfections of PU-learning can cause the set $\mathcal{D}_{\text{obs}}$ to contain samples from both correct observed-novel categories and incorrect source categories. Although previous research has addressed the noisy label problem~\cite{tanaka2018joint, song2022learning}, these studies focused on noisy labels of source categories in a closed setting. Consequently, future Open-LACU research should prioritize further developing PU-learning models, particularly within the context of neural networks. However, these developments should be done in conjunction with the uncharted territory of noisy labels associated with an augmented $K + 1$'th background category. 

It is also essential to acknowledge that the results of the CIFAR10 experiment displayed considerable variability from one epoch to the next. Although we presented the top-performing model over the last 50 epochs, we must admit that these classifiers exhibited fluctuations in results, alternating between accurately generalising the source categories and accurately generalising the augmented $K + 1$'th background category. This inconsistency can be directly attributed to the imperfect nature of $\mathcal{D}_{\text{obs}}$. Then, bar PU-learning, it will also be intriguing to determine which SSL models are better suited for handling an augmented $K + 1$'th background category. Based on our observations, we hypothesise that SSL models that make use of guessed labels (such as MixMatch~\cite{berthelot2019mixmatch}, Mean-teacher~\cite{tarvainen2017mean}, pseudo-labelling~\cite{lee2013pseudo}, and meta pseudo-labelling~\cite{pham2021meta}) might not be as effective in the Open-LACU setting as intrinsic SSL models.

The Open-LACU problem can be framed as an imbalanced class distribution problem~\cite{vuttipittayamongkol2021class}. Generally, imbalanced class distributions refers to the number of labelled samples being skewed towards certain categories. However, in our case, the imbalance refers to the number of categories associated within a single label. Specifically, the augmented $K + 1$'th background category will be skewed since it encapsulates many observed-novel categories in the domain. With regard to SSL, methods that rely on estimated labels can readily gravitate towards the incorrectly guessing the $K + 1$'th label for source categories, especially considering that this skewed category also includes noisy labels. Therefore, in the realm of Open-LACU, future studies should also consider the imbalanced problem but, again, for a the augmented $K + 1$'th background category instead of the source categories. 

Finally, given that GANs have been successfully employed in PU-learning~\cite{chiaroni2020counter, ijcai2018312, hu2021predictive}, SSLs~\cite{NIPS2016_8a3363ab, dai2017good, kumar2017semi, li2020semi}, and OSR~\cite{ge2017generative, neal2018open, jo2018open, chen2021adversarial}, we advocate for the development of an Open-LACU GAN. More specifically, we propose the development of an end-to-end training scheme to achieve Open-LACU whereby 1) the PU-learning step is integrated with a GAN that exclusively generates samples from observed-novel categories (as developed by Chiaroni et al.~\cite{chiaroni2020counter}), 2) the generator is applied in conjunction to a discriminator that is designed as an SSL model~\cite{dai2017good}, which 3) has been shown effective in OSR~\cite{engelbrecht2023link}. Such an Open-LACU GAN would represent a promising avenue to address the challenges posed by skewed label distributions and noisy labels, while also streamlining the training process.

\section{Conclusion}
In this study, we introduced a groundbreaking learning policy that effectively distinguishes between observed-novel categories and unobserved-novel categories. Open-set learning with an augmented category by exploiting unlabelled data, abbreviated as Open-LACU, requires models to generalise across three distinct category types: 1) the $K$ source categories, 2) an additional augmented $K + 1$'th background category representing novel categories observed in the unlabelled training data, and 3) another augmented $K + 2$'nd unknown category for novel categories unobserved during training. By enabling neural network classifiers to differentiate between these diverse category types, we enhance the versatility and performance of classifier models.

After presenting this learning policy, we also proposed the first Open-LACU model that leverages a neural network classifier with $K + 1$ output nodes and a reject option to accommodate the $K + 2$ category. Our results are not only state-of-the-art but also the first-of-their-kind in this emerging field. However, the Open-LACU problem is intricate and intersects with various research domains, making it a complex challenge. Therefore, to advance Open-LACU further, a holistic approach that consolidates the insights from our experiments in the diverse research areas is a promising direction for future development.

\clearpage
\section*{Declaration of Generative AI and AI-assisted technologies in the writing process}
During the preparation of this work the author(s) used Grammarly and ChatGPT in order for grammatical improvement for better readability of the study. After using this tool/service, the author(s) reviewed and edited the content as needed and take(s) full responsibility for the content of the publication.

\printbibliography
  
\end{document}